\documentclass[letterpaper, 10 pt, conference]{ieeeconf}  

\IEEEoverridecommandlockouts                              

\usepackage{amsmath} 
\usepackage{amssymb}  
\usepackage{subcaption}
\usepackage{graphicx}
\usepackage{hyperref}

\DeclareMathOperator*{\argmin}{arg\,min}

\title{\LARGE \bf
Neural Field Representations of Articulated Objects for Robotic Manipulation Planning
}

\author{Phillip Grote$^{1}$, Joaquim Ortiz-Haro$^{1}$, Marc Toussaint$^{1}$, Ozgur S. Oguz$^{2}$
\thanks{$^{1}$TU Berlin, Germany, $^{2}$Bilkent University, Türkiye. This work was supported by TUBITAK under 2232 program with project number 121C148 (``LiRA``), and the German-Israeli Foundation for
Scientific Research (GIF) grant I-1491-407.6/2019. Joaquim Ortiz-Haro
thanks the International Max-Planck Research School for
Intelligent Systems (IMPRS-IS) for the support.
}%
}

\begin{document}

\maketitle
\thispagestyle{empty}
\pagestyle{empty}

\begin{abstract}

Traditional approaches for manipulation planning rely on an explicit geometric model of the environment to formulate a given task as an optimization problem. However, inferring an accurate model from raw sensor input is a hard problem in itself, in particular for articulated objects (e.g., closets, drawers). In this paper, we propose a \emph{Neural Field Representation} (NFR) of articulated objects that enables manipulation planning  directly from images. 
Specifically, after taking a few pictures of a new articulated object, we can forward simulate its possible movements, and, therefore, use this neural model directly for planning with trajectory optimization. Additionally, 
this representation can be used for shape reconstruction, semantic segmentation and image rendering, which provides a strong supervision signal during training and generalization. 

We show that our model, which was trained only on synthetic images, is able to extract a meaningful representation for unseen objects of the same class, both in simulation and with real images. Furthermore, we demonstrate that the representation enables robotic manipulation of an articulated object in the real world directly from images.

\noindent Video: \href{https://phgrote.github.io/nfr/}{https://phgrote.github.io/nfr/}

\end{abstract}

\begin{figure*}
     \centering
       \begin{subfigure}[b]{0.19\textwidth}
         \centering
         \includegraphics[width=\textwidth]{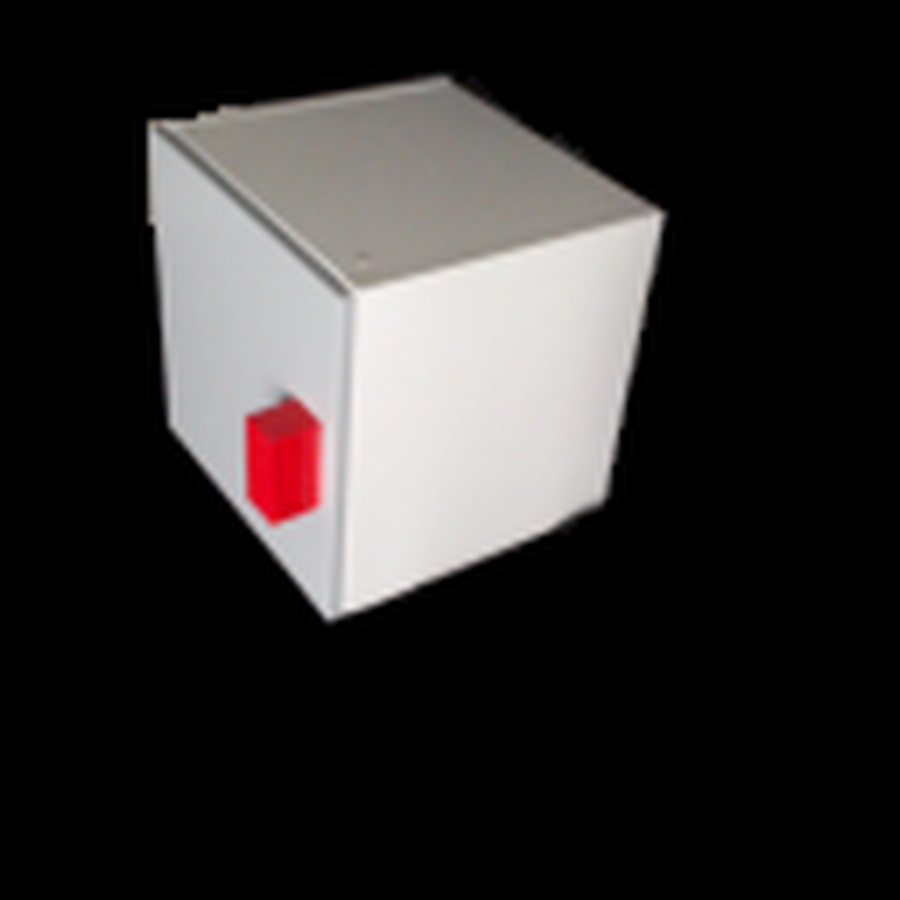}
         \caption{See}
         \label{fig:pipeline_see}
     \end{subfigure}
     \begin{subfigure}[b]{0.19\textwidth}
         \centering
         \includegraphics[width=\textwidth]{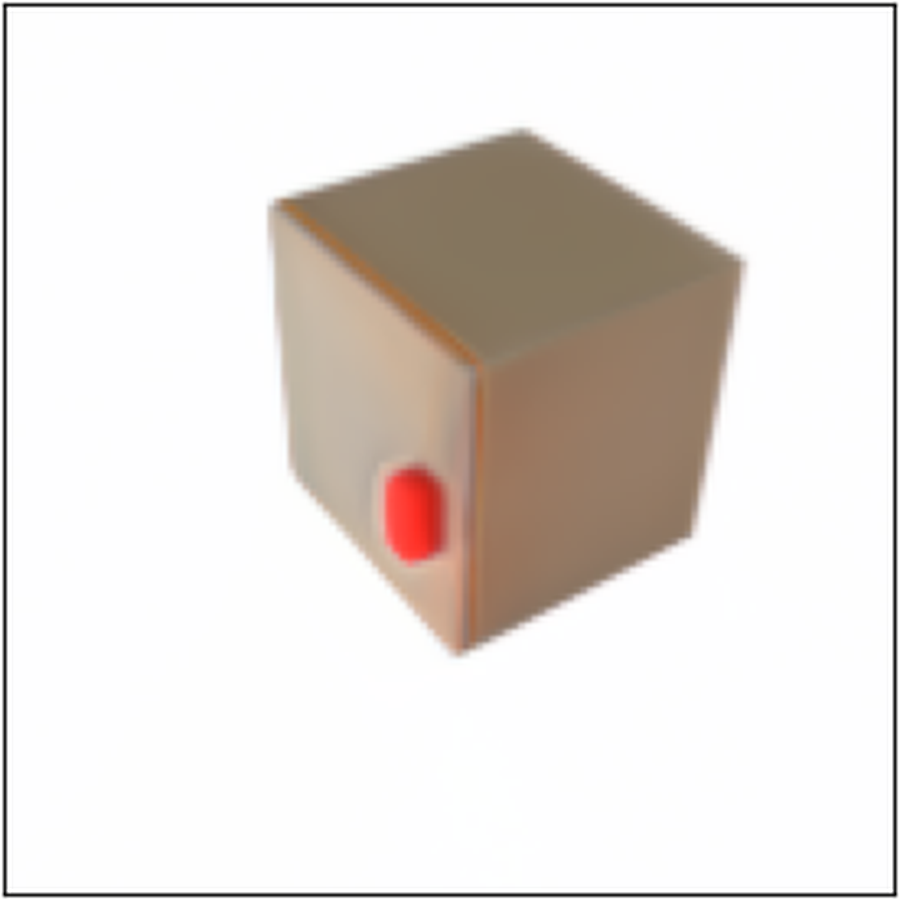}
         \caption{Find $\mathbf{z}$}
         \label{fig:pipeline_z}
     \end{subfigure}
        \begin{subfigure}[b]{0.19\textwidth}
         \centering
         \includegraphics[width=\textwidth]{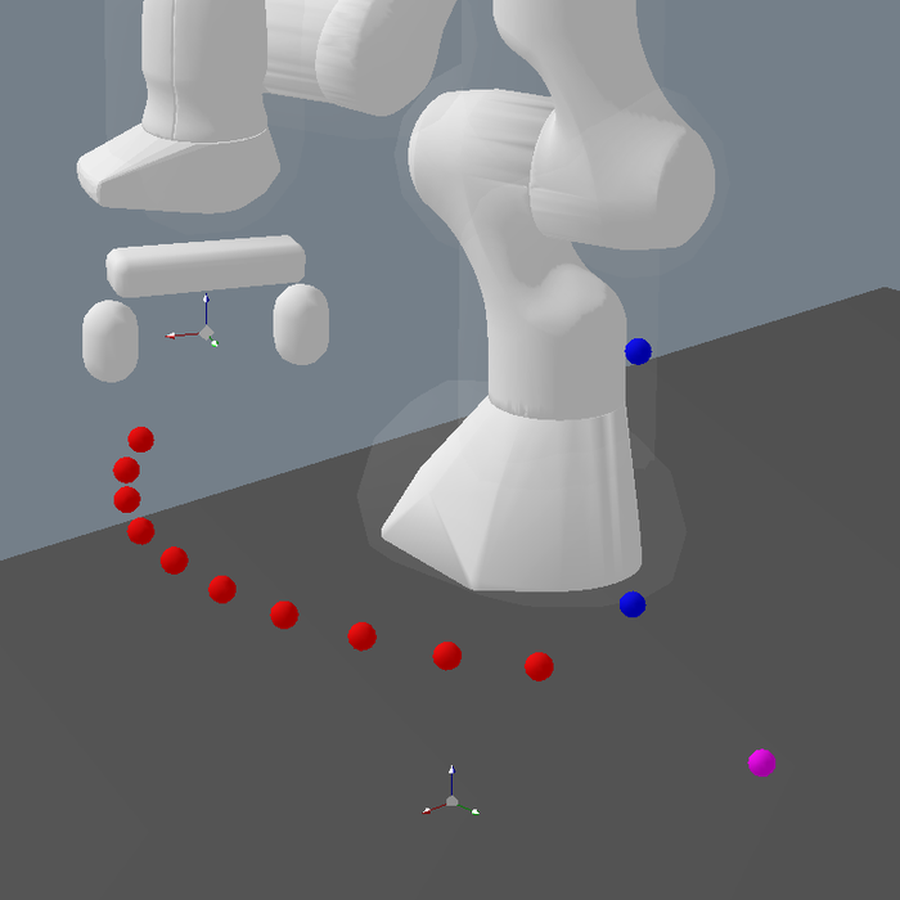}
         \caption{Compute $\mathbf{k}$}
         \label{fig:pipeline_kps}
     \end{subfigure}
     \begin{subfigure}[b]{0.19\textwidth}
         \centering
         \includegraphics[width=\textwidth]{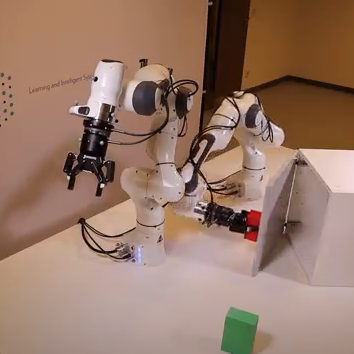}
         \caption{Open}
         \label{fig:pipeline_open}
     \end{subfigure}
     \begin{subfigure}[b]{0.19\textwidth}
         \centering
         \includegraphics[width=\textwidth]{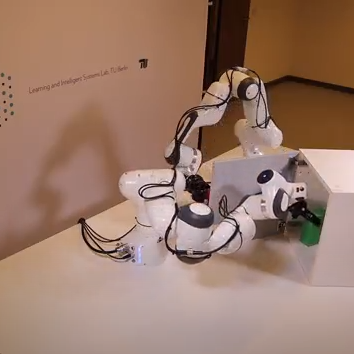}
         \caption{Place}
         \label{fig:pipeline_place}
     \end{subfigure}

        \caption{Interaction with articulated objects: First, the robot observes a new object (a); The latent code $\mathbf{z}$ is found by minimizing the image loss between the observed real images and the generated images (b); We predict keypoints $\mathbf{k}$ by forward simulating the motion (c); Finally, all keypoints are used to formulate an optimization problem (d) and (e).}
        \label{fig:pipeline}
\end{figure*}

\section{INTRODUCTION}

Robots could support humans with everyday chores like cleaning if they were able to reliably interact with articulated objects such as closets and drawers. Every concrete interaction with the environment (e.g., the opening of a closet) can be formalized as a constrained minimization problem. By defining the objective function in terms of manipulation features, which map the environment to numerical quantities (e.g., the position of an object), we are not limited to solve only for the robot's own movement, but are able to optimize for instance the location of other objects within the environment. In order to formulate such optimization problems the robot needs a good representation of objects in the scene. In general, this representation has to be inferred from raw sensory inputs like images or point clouds. 

Traditional approaches represent objects explicitly, for instance as a mesh or a combination of geometric shapes (e.g., spheres, boxes, etc.). The dynamic behavior of articulated objects is modeled explicitly as well, e.g., by inferring the location of the rotational axes for revolute joints~\cite{klingbeil2010learning,sturm2011probabilistic,jain2021screwnet} or by estimating how the perceived object relates to a known canonical representation~\cite{li2020category} or prototype~\cite{englert2018kinematic}. Similar to the work of Eisner et al.~\cite{EisnerZhang2022FLOW}, we investigate the use of implicit representations for articulated objects, demonstrate how such representations can be inferred from raw RGB images, and how they can be used for manipulation planning.

An implicit neural field representation can be inferred from raw sensory RGB input by minimizing the loss between rendered and observed images, thereby making depth sensors, traditional approaches rely on, dispensable. We encode this representation by a low dimensional structured latent code. The proposed structure of the latent code allows us to manipulate the latent code in a predictable way in order to \emph{simulate} the whole range of motion of a perceived object. Finally, we show that this representation can be  transformed to a semantic 3D keypoint representation~\cite{manuelli2019kpam} to enable category-level manipulation using existing manipulation planning frameworks~\cite{toussaint2015logic}. The proposed interaction with an articulated object is depicted in Fig.~\ref{fig:pipeline}.    

To summarize, our main contributions in this work are: 
\begin{itemize}
    \item Framework for generating neural field representations conditioned on a structured latent code, which enables the forward simulation of possible movements 
    \item Integrated architecture to extract implicit object representations from posed images, in order to generate images, semantically labeled point clouds and keypoint predictions for arbitrary articulations
    \item Integration of the neural representation within a sequential manipulation planning framework
\end{itemize}

We evaluate our approach in multiple ways. First, we demonstrate the generative capabilities by interpolating between different representations and by generating new representations for unobserved articulations. Next, we evaluate the prediction of keypoint positions, which is essential for manipulation planning. We demonstrate in simulation as well as on a real robot that we are able to manipulate an articulated object based on the representation extracted from posed images. Finally, we show that our method is robust to out-of-distribution scenarios, i.e., it can infer good representations from real RGB images, even though our architecture was trained on synthetic images with different camera parameters.   


\section{RELATED WORK}

\subsection{Implicit Representations in Robotics}
Implicit representations are gaining popularity within the robotics community. They have been used for long-horizon planning from visual inputs ~\cite{ha2022deep}, navigation ~\cite{adamkiewicz2022vision}, pose estimation ~\cite{yen2021inerf} and reinforcement learning ~\cite{driess2022reinforcement}. Furthermore, they are capable of predicting how articulated parts move under kinematic constraints without knowing the explicit kinematic model ~\cite{EisnerZhang2022FLOW}.  

Instead of adopting NeRF as in~\cite{adamkiewicz2022vision,yen2021inerf,driess2022reinforcement}, we are using \emph{Scene Representation Networks} (SRN)~\cite{sitzmann2019scene} as an underlying implicit representation in order to encode surface distances directly. By adopting an auto-decoder approach instead of encoding observations directly~\cite{driess2022reinforcement,ha2022deep} we are robust to out-of-distribution scenarios. Instead of using a static representation for pose estimation
~\cite{adamkiewicz2022vision, yen2021inerf}, this work focuses on how to manipulate an inferred representation in order to predict how the articulation of the object affects the position of keypoints. Finally, by predicting the handle position our method does not require a suction-type gripper as in \cite{EisnerZhang2022FLOW}.

\subsection{Implicit Representations for articulated objects}
Representing 3D objects as continuous and differentiable implicit functions 
is a well established field of research~\cite{genova2020local,jiang2020local,mildenhall2021nerf,niemeyer2020differentiable,park2019deepsdf,saito2019pifu,sitzmann2019scene,trevithick2021grf,xu2019disn}. This line of research typically focuses on static objects, but representing dynamic articulated objects is starting to emerge as a new direction~\cite{jiang2022ditto,mu2021sdf,aNerf,tseng2022cla}. 

Mu et al.~\cite{mu2021sdf} propose to use an \emph{Articulated-Sign Distance Function} (A-SDF), a learned \emph{Sign Distance Function} (SDF) based on the work of Park et al.~\cite{park2019deepsdf}, to represent articulated objects. In regard to the separation of shape code and articulation code, our approach is similar, but instead of using an SDF as an implicit function we are using a more general function $\Phi$, which maps spatial coordinates to feature vectors. This allows us to formulate the reconstruction loss on images rather than point clouds. Thus we do not have to assume that point cloud data is available.

Su et al. \cite{aNerf} extend NeRF~\cite{mildenhall2021nerf} for learning a 3D representation of the human body from 2D observations. While they refine an initial estimation of the articulation given by an off-the-shelf estimator, we are estimating the articulation directly without the need of an additional estimator. 

Learning the motion constraints through interaction is addressed by \cite{jiang2022ditto}. Our approach does not require additional interaction with a new instance from a learned category in order to perform motion planning.

The study by Tseng et al.~\cite{tseng2022cla} addresses articulated objects by extracting an explicit kinematic model of the perceived object by fitting a rotation axis between intersecting parts. In contrast, our approach directly generates keypoint representations for different articulations in order to perform motion planning.

\subsection{Articulated Objects}
The manipulation of articulated objects is a well known problem. In order to enable robotic manipulation it has been proposed to extract an explicit kinematic model from demonstration either using fiduciary markers~\cite{sturm2011probabilistic,niekum2015online} or by tracking features within the observation~\cite{pillai2015learning,jain2020learning}. Others proposed to extract an explicit kinematic model through interactive perception~\cite{katz2008manipulating, hausman2015active, martin2016integrated, martin2022coupled, martin2014online}. Another line
of research assumes knowledge of the kinematic structure of a broader category and only adjusts the parametrization to the observed instance from observed depth data~\cite{abbatematteo2019learning, li2020category}. 

Instead of extracting an explicit model of the kinematic structure we are using an implicit representation. We are able to infer representations from posed RGB images and do not require depth data. Furthermore, our method does not require any interactions with the perceived object in order to construct a model of its kinematic structure.






\section{BACKGROUND}

Our approach extends 
\emph{Scene Representations Networks} \cite{sitzmann2019scene} for manipulation planning of articulated objects. 
We first summarize the original framework, and present our contributions and extensions in Sec. \ref{sec:nfr}.

We represent objects implicitly with a neural field:  a function $\Phi_{\theta}  \in \mathcal{X}$, which maps 3D spatial coordinates $\mathbf{x} $ to $n$-dimensional feature vectors $\mathbf{v}$,
\begin{align}
    \Phi_{\theta}: \mathbb{R}^3 \rightarrow \mathbb{R}^n, ~\mathbf{x}\mapsto \Phi_{\theta}(\mathbf{x}) = \mathbf{v}.
\end{align}
This function is implemented with a neural network parameterized by the weight vector $\theta \in \mathbb{R}^l$. 
Given $\Phi_{\theta}$, we can render images using a differentiable rendering algorithm $\Theta_{\psi}$, for any camera extrinsic $\mathbf{E}$ and intrinsic $\mathbf{K}$ parameters:
\begin{equation}
\begin{aligned} 
    \Theta_{\psi}: \mathcal{X} \times \mathbb{R}^{3\times4} \times \mathbb{R}^{3\times3} &\rightarrow \mathbb{R}^{H\times W \times 3},\\
    (\Phi, \mathbf{E}, \mathbf{K}) &\mapsto \Theta(\Phi,\mathbf{E}, \mathbf{K}) = \hat{\mathcal{I}}.
\end{aligned}
\end{equation}
We generate images by mapping the feature vectors at all surface points to their corresponding RGB values. Surface points are obtained by querying $\Phi_{\theta}$ repeatedly and mapping the corresponding feature vectors to step sizes along camera rays (differentiable raymarching). $\Theta_\psi$ is implemented by multiple neural networks and we collect all weights in one weight vector $\psi$. 

By using a hypernetwork $H_\phi$~\cite{ha2016hypernetworks} it is possible to find a $k$-dimensional subspace of the space of weights of the neural vector field  $\theta \in \mathbb{R}^{l}$, which allows to represent objects with a $k-$dimensional latent code $\mathbf{z}$
\begin{align}
    H_\phi: \mathbb{R}^k \rightarrow \mathbb{R}^{l},~ \mathbf{z} \mapsto H_{\phi}(\mathbf{z}) = \theta~,
\end{align}
with $k < l$, which suffices to represent all instances of a certain object class $\mathcal{K} \subset \mathcal{X}$~\cite{sitzmann2019scene}. By implementing $H_\phi$ as a neural network and optimizing the weight vector $\phi$, we are learning a suitable prior of 3D surfaces. This prior is necessary to estimate a plausible 3D surface shape given a (possibly small) set of 2D images~\cite{xie2022neural}. 

Given a set of posed images $\mathcal{D} = \{ \mathcal{C}_j \}_{j=1}^{N_\text{obj}}$ with $\mathcal{C}_j := \{(\mathcal{I}_{j,i}, \mathbf{E}_{j,i}, \mathbf{K}_{j,i})\}_{i=1}^{N_\text{view}}$ of different objects, 
we can learn to represent objects of a given class, by training the latent codes and all other weights jointly using the auto-decoder framework:
\begin{align}
    \argmin_{\mathbf{z}_j, \phi, \psi} \sum_{j=1}^{N_\text{obj}}\sum_{i=1}^{N_\text{view}} ||\Theta(\Phi_{H(\mathbf{z}_j; \phi)}, \mathbf{E}_i, \mathbf{K}_i; \psi) - \mathcal{I}_i||_2^2. \label{eq:nfr_min}
\end{align}
The latent code $\mathbf{z}_\text{new}$ for a previously unseen instance $\mathcal{C} := \{(\mathcal{I}_{i}, \mathbf{E}_{i}, \mathbf{K}_{i})\}_{i=1}^{N_\text{view}}$ is obtained through optimization as well:
\begin{align}
    \argmin_{\mathbf{z}_\text{new}} \sum_{i=1}^{N_\text{view}} |\!| \Theta(\Phi_{H(\mathbf{z}_\text{new}; 
    \phi)}, \mathbf{E}_i, \mathbf{K}_i; \psi) - \mathcal{I}_i|\!|_2^2 \label{eq:nfr_min2}.
\end{align}

Because the latent code $\mathbf{z}$ is not generated by encoding the observations but is found through optimization instead, this approach is referred to as an auto-decoder framework~\cite{bojanowski2017optimizing,park2019deepsdf}. Due to this additional optimization step, auto-decoding is slower than a feed forward encoder approach. However, auto-decoding is more robust in certain out-of-distribution scenarios~\cite{xie2022neural}. For instance, auto-decoding is able to infer good latent codes with low reconstruction loss even if the camera poses of the observations were not seen during training~\cite{sitzmann2019scene}. These benefits have contributed to the wide adoption of the auto-decoder approach~\cite{jang2021codenerf,liu2021editing,park2019deepsdf,ramon2021h3d,sitzmann2021light,sitzmann2019scene,tretschk2021non,yang2021deep}. 

\section{NEURAL SCENE REPRESENTATIONS FOR ARTICULATED OBJECTS}\label{sec:nfr}

For manipulation planning of articulated objects, we propose a method that can forward simulate the possible motions of a given object. 
\begin{figure}[t]
\centering
\includegraphics[width=.45\textwidth]{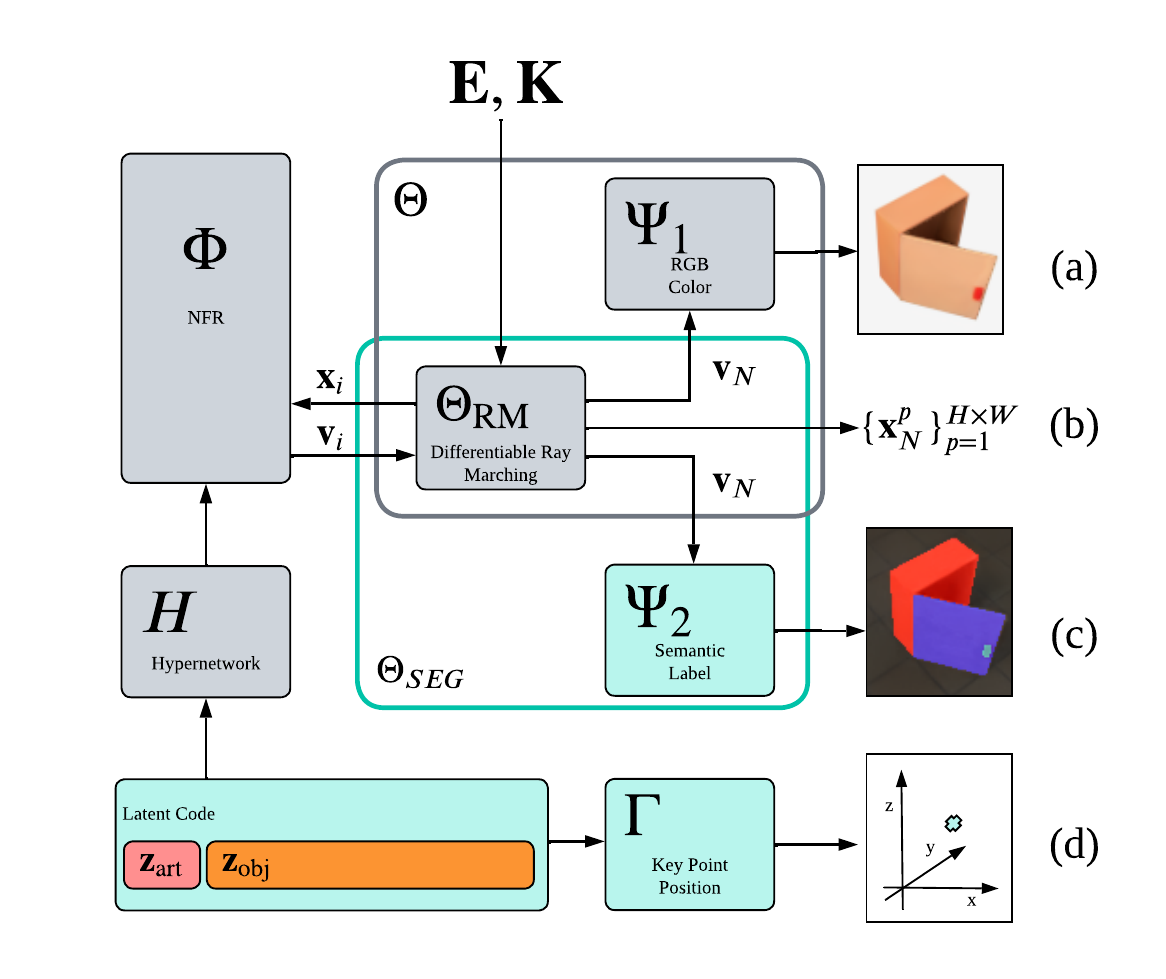}
\caption[Overview of the prediction scheme with generated outputs]{Overview: latent code $\mathbf{z}$ is mapped via $H$ to $\Phi$, which is queried repeatedly in order to extract surface points through differentiable raymarching (b); Feature vectors at surface points are mapped to RGB values to generate  RGB images (a) and to semantic labels (c); 3D positions of keypoints (e.g. handle) are directly obtained from $\mathbf{z}$ (d).}
\label{fig:nfr_overview}
\end{figure}
With our extensions to SRNs \cite{sitzmann2019scene}, namely structured latent code, keypoint prediction and semantic labeling, we obtain a novel architecture which enables the desired forward simulation of motion (Fig. \ref{fig:nfr_overview}). Furthermore, by forward simulating the motion of the object and predicting keypoints for arbitrary articulations we are able to perform manipulation planning.


In the following sections we will explain our extensions, the training of the whole model and how previously unseen objects are handled.


\subsection{Latent code}\label{sec:emb}
We define the latent code of the object instance as, \begin{align}
   \mathbf{z} := \begin{bmatrix}
    \mathbf{z}_\text{art} \\
    \mathbf{z}_\text{obj}
    \end{bmatrix}.
\end{align}
The latent code comprises two distinct parts: the articulation code $\mathbf{z}_\text{art}$ and the object code $\mathbf{z}_\text{obj}$. The articulation code encodes the articulation and the object code encodes the shape and the appearance of the object. 

Instead of mapping the whole range of a single joint to the interval $I := [0,1]$ we are using a two dimensional representation $\mathbf{z}_\text{art} \in \mathbb{R}^2$. This allows us to introduce a normalization layer to transform $\mathbf{z}_\text{art}$. The transformed articulation code $\mathbf{\hat{z}_\text{art}}$ lies on a unit circle and in order to avoid discontinuities one half of the unit circle represents all possible articulations, while the second half of the unit circle mirrors the first. With the proposed normalization we ensure a uniform distribution of $\hat{\mathbf{z}}_\text{art}$, even though we are using a gaussian prior on $\mathbf{z}_\text{art}$. The proposed parameterization was motivated by the work of Salimans and Kingma~\cite{salimans2016weight} for improving the speed of convergence.

\subsection{Forward Simulation by Latent Code Manipulation} After training, when we see a new object instance we first optimize the complete latent code by minimizing the image reconstruction loss. Now, we can simulate the movement by modifying the articulation code, while keeping the object code constant. For each new code, we can simulate the movement by generating images, segmentation masks and predict keypoint positions. Finally, the information generated by simulating the movement is used for manipulation planning with trajectory optimization (Sec. \ref{sec:planning}).

\subsection{Semantic Segmentation}
Using differentiable raymarching we are able to generate a multiset of feature vectors $\mathcal{V} = \{ \mathbf{v}_N^p \}_{p=1}^{H \times W}$ (Fig. \ref{fig:nfr_overview}b). These  feature vectors can be mapped to RGB colors via $\Psi_1$ (Fig. \ref{fig:nfr_overview}a) or to semantic labels via $\Psi_2$ (Fig. \ref{fig:nfr_overview}c). Thus, we are able to generate semantically labeled point clouds of the object from arbitrary viewpoints ($\mathbf{E}$, $\mathbf{K}$) \emph{and} arbitrary articulations.


\subsection{Keypoint Prediction}
A latent code representation $\mathbf{z}$ can be used to predict the 3D positions of specific keypoints (Fig. \ref{fig:nfr_overview}d). For instance, on our closet dataset we defined the center of the handle, the hinge joints and a goal location inside the closet as keypoints. In order to predict the keypoint positions for an arbitrary articulation $q$, we first generate a new latent code $\mathbf{z}_\text{new}$ and then map this generated latent code $\mathbf{z}_\text{new}$, via the neural network $\Gamma_\gamma$, to the predicted keypoint positions $\mathbf{k}_\text{new}$.

\subsection{Training}\label{sec:train}
Here, we describe the training of our framework on closets, including the data generation process and the loss function used.

\subsubsection{Data Generation}\label{sec:data_gen}
We generated a dataset containing $N_\text{closet} = 1000$ closet models with varying shapes and appearances. For each closet model we generated $N_\text{art} = 100$ uniformly distributed articulations of the door between $0^\circ$ ($q=0)$ and $90^\circ$ ($q=1$). Thus in total our dataset $\mathcal{D}$ is composed of $M = N_\text{closet} \cdot N_\text{art} = 100000$ instances. For each instance we generated $N_\text{view}= 10$ posed images with a resolution of $128 \times 128$ using NViSII~\cite{morrical2021nvisii}, a scriptable tool for photorealistic image generation. Additionally, we varied the lighting conditions. For each instance we generated the ground truth position of the handle, the hinges and the goal location inside the closet.

\subsubsection{Loss Function}
We are optimizing all object codes $\{\mathbf{z}^l_\text{obj}\}_{l=1}^{N_\text{closet}}$ and the weights of all networks ($\phi$: hypernetwork; $\gamma$: keypoint prediction; $\theta_{\text{RM}}$: raymarching; $\psi_1$: RGB rendering; $\psi_2$: semantic labelling) jointly,
\begin{equation}\label{loss}
\begin{aligned}
\argmin_{\substack{\{\mathbf{z}^l_\text{obj}\}_{l=1}^{N_\text{closet}}\\
\phi, \gamma\\
\theta_\text{RM}, \psi_1,\psi_2}} \sum_{l = 1}^{N_\text{closet}} \sum_{k = 1}^{N_\text{art}} \sum_{i = 1}^{N_\text{view}} &\mathcal{L}_\text{SRN} + \lambda_1\mathcal{L}_{\text{SEG}} + \lambda_2\mathcal{L}_{\text{KP}}.
\end{aligned}
\end{equation}
In this formulation $\mathcal{L}_\text{SRN}$ is comprised of the image loss ($\mathcal{L}_\text{img}$) and two regularization terms, for more details please refer to~\cite{sitzmann2019scene}. The other loss components are defined as follows:
\begin{align*}
    \mathcal{L}_\text{SEG} &= \text{CE}(\Theta_\text{SEG}(\Phi_{H(\mathbf{z}; \phi)}, \mathbf{E}_i^{l,k}, \mathbf{K}_i^{l,k}; \theta_\text{RM}, \psi_2), \mathcal{J}_i^{l,k}),\\[.25cm]
    \mathcal{L}_\text{KP} &= || \Gamma(\mathbf{z}; \gamma) - \mathcal{P}^{l,k}||_2^2,
\end{align*}
where $\text{CE}(\cdot)$ is the cross entropy loss between the predicted segmentation generated via $\Theta_\text{SEG}$ and the ground-truth segmentation $\mathcal{J}_i^{l,k}$. 
During training, the ground truth articulation codes $q^{l,k}$ are used. The ground-truth keypoint positions $\mathcal{P}^{l,k}$ and images $\mathcal{I}_i^{l,k}$ are provided during training as well.
The weights $\lambda_{1}$ and $\lambda_2$ control the relative importance of each loss term during training.

\begin{figure}[!h]
\centering
\includegraphics[width=.16\textwidth]{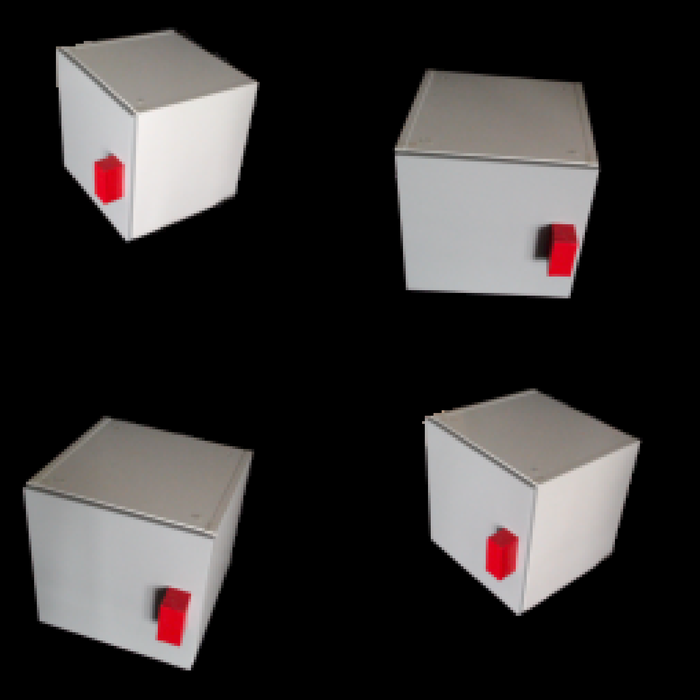}\hfill
\includegraphics[width=.16\textwidth]{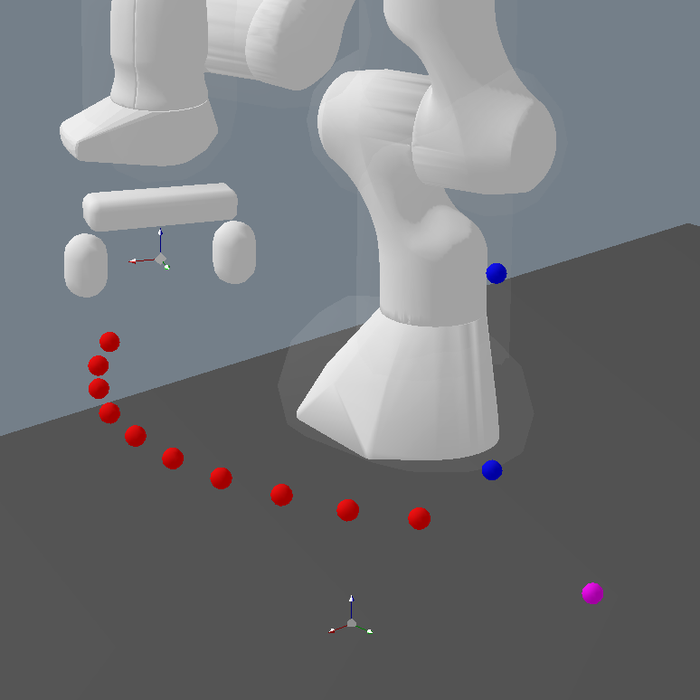}\hfill
\includegraphics[width=.16\textwidth]{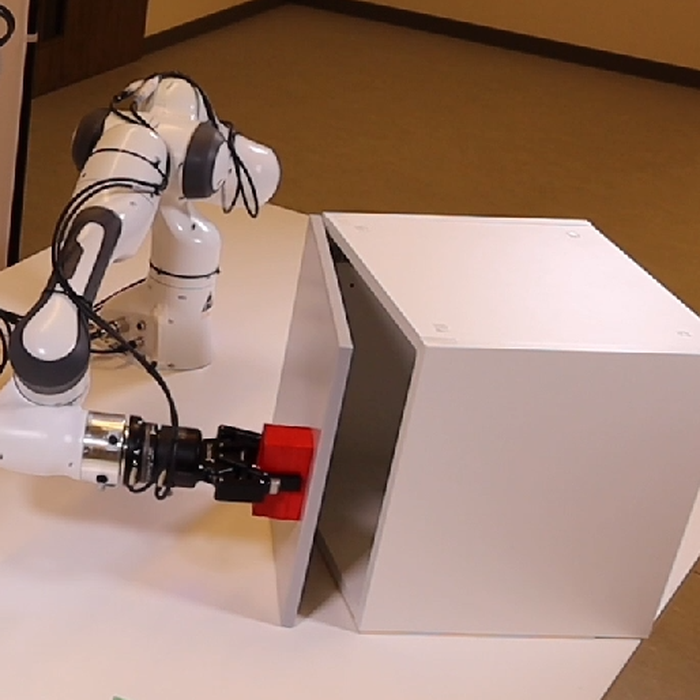}
\caption{Interaction with a closet: Four input images (left); Simulated motion: handle positions in red, hinge positions in blue and goal location in magenta (middle); final interaction on a real robot (right).}
\label{fig:tamp}
\end{figure}

\subsection{Inference}
Given a trained model and a set of images of a previously unseen articulated object, we are able to find the corresponding latent code of the object by minimizing the image loss. In contrast to the training phase, the weight vector of all neural networks are kept constant. Additionally, since we do not have access to the ground truth semantic segmentation and the positions of the keypoints, we set $\lambda_1 = \lambda_2 = 0$.

\begin{figure*}[!h]
    \centering
        \includegraphics[width=.6\textwidth]{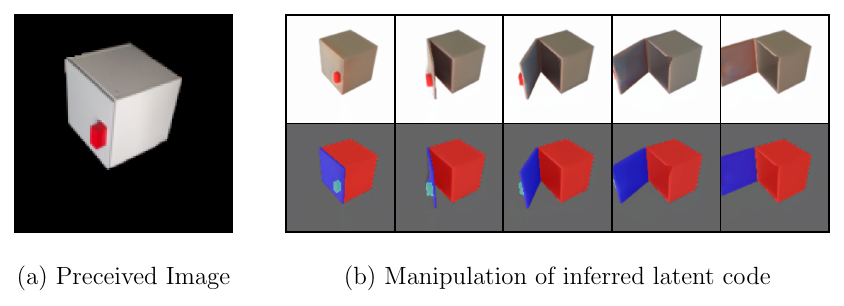}
    \caption{Forward simulation of motion: (a) shows one input image of the perceived object; The top row in (b) shows the RGB renderings generated with $\Theta$, whereas the bottom row depicts the semantic segmentation generated by $\Theta_\text{SEG}$.}
    \label{fig:emb_manipulation}
\end{figure*}
\section{MANIPULATION PLANNING WITH NEURAL REPRESENTATIONS}\label{sec:planning}
In order to perform manipulation planning we integrated our neural field representation of articulated objects with the constraint-based trajectory optimization and manipulation planning framework used within \emph{Logic-Geometric Programming} (LGP)~\cite{toussaint2015logic}.
With this framework our method works as follows: 
\begin{enumerate}
\item The robot takes a few pictures of an unseen closet.

\item The latent code that corresponds to the closet is computed by minimizing the image reconstruction loss. 

\item Movement of the closet is simulated by interpolating the articulation component of the latent code, from the estimated current value to a desired value. During this forward-simulation of the neural model, the trajectory of a set of keypoints is predicted and stored. 

\item The predicted keypoint trajectory is used to define a trajectory optimization problem.

\item The optimization problem is solved with constrained  optimization, and the robots execute the resulting motion.

\end{enumerate}

By predicting these keypoint positions, the motion constraints of the object are considered. As depicted in Fig. \ref{fig:tamp}, for each articulation the position of the hinges and target location remain constant, while the handle moves along an object specific trajectory. By mapping consecutive articulations to corresponding time steps we can define different tasks such as the \emph{opening} or \emph{closing} of a closet. 
Specifically, the interaction with the object is discretized into $T \in \mathbb{N}$ steps. Given an inferred articulation code and a target, we can map each intermediate step to a specific articulation using a linear interpolation in latent space. By mapping the $T$ latent codes (combination of interpolated articulation and inferred object codes) to keypoint positions, we are able to formulate a constrained minimization problem.

\section{EXPERIMENTAL EVALUATION\label{cha:eval}}
We evaluate our framework in multiple ways. First, we evaluate the ability of our learned model to render images. Next, we demonstrate that given a latent code representation, the motion of an articulated object can be simulated.
Furthermore, we evaluate the keypoint estimation of an observed object as well as the keypoint prediction for arbitrary articulations. Finally, we demonstrate the robotic manipulation of different object classes in simulation and on a real robot. 

Our model was trained on the training dataset described in Sec. \ref{sec:train}. For our evaluation we use two different datasets:
$\mathcal{D}_\text{SYNT}$ and $\mathcal{D}_\text{REAL}$. Both datasets have $N_\text{view} = 10$ views of each particular instance. $\mathcal{D}_\text{SYNT}$ was generated similar to the training dataset. For $\mathcal{D}_\text{REAL}$ we manually collected posed images from a single (real) closet.

\subsection{Image Rendering}

\begin{figure}[!ht]
    \centering
    \includegraphics[width=.45\textwidth]{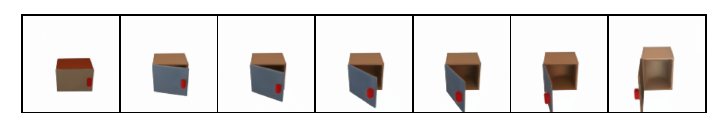}
    \caption{Interpolation of latent codes across articulation code and object code: the latent codes corresponding to the left- and rightmost image were found through optimization.}
    \label{fig:interpolation}
\end{figure}

Latent code representations are found through the minimization of the image loss between the observed image and the image rendered by our trained model. Fig.~\ref{fig:interpolation} confirms that for our trained model interpolations between latent codes correspond to semantically meaningful and smooth interpolations in image space, which is required in order to find good latent representations for a broad range of objects. 
Furthermore, our framework is also able to find good latent code representations for real images and we can simulate the whole range of motion of the perceived object (Fig.~\ref{fig:emb_manipulation}). 

The ability to interpolate between latent codes, the generalization to real images and the ability to simulate the motion confirms that we have learned a strong prior for the given object category.

\subsection{Keypoint Estimation and Forward Simulation of Motion}

In this section we evaluate the keypoint estimation and prediction. First, we describe the baseline used. Next, we evaluate the keypoint estimation of the observed object and the keypoint prediction for arbitrary articulations.  

\subsubsection{Baseline}
As a baseline we trained a standard image encoder $\mathcal{E}$ similar to the one used in~\cite{ha2022deep}, which adopted the U-net architecture~\cite{ronneberger2015u} with ResNet-34 ~\cite{he2016deep} as its downward path. Each image $\mathcal{I}_i$ together with its pose $\mathbf{E}_i$ with $i \in \{1, ...,N_\text{view}\}$ of a single instance is encoded. The final latent code $\mathbf{z}_\text{ResNet}$ is obtained by taking the average of all $N_\text{view}$ image encodings. 
The neural network $\Gamma_\text{ResNet}$ maps latent codes to keypoint positions, \emph{and} estimates the current articulation $q$ of the perceived object explicitly. 

In contrast to our approach, the baseline implementation is not capable of generating new representations for different articulations of the observed object and thus can only infer features, e.g., keypoints, for the perceived object. In order to compare the baseline to our model we are required to provide additional knowledge about the geometric properties and behavior of any given object. For example, on our closet dataset we assume a vertical axis of rotation at the hinge position. Only with this additional assumption we can predict the positions for different articulations by the baseline model.

\subsubsection{Keypoint Estimation of Observed Configurations}\label{sec:kps_eval}
In this section we evaluate the estimated keypoint position for observed objects. We compare our implementations, with and without articulation code normalization (Sec.~\ref{sec:emb}), and the ResNet baseline. 
\begin{figure}[h!]
    \centering
    \includegraphics[width=.45\textwidth]{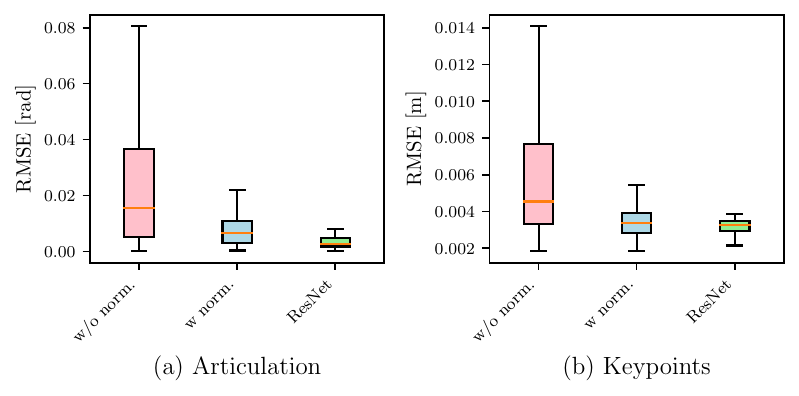}
    \caption{Articulation and keypoint estimation: (a) shows the prediction error on the estimated articulation angle; (b) shows the RMSE error on the predicted keypoint positions}
    \label{fig:eval1}
\end{figure}
Our results in Fig.~\ref{fig:eval1} show that with the proposed normalization of the articulation code we achieve comparable results to a classic image encoder. All methods achieve subcentimeter accuracy, while our methods provides additional benefits like generating point clouds with semantic annotations and generating estimates for arbitrary articulations for objects with unknown dynamic behavior.

\subsubsection{Forward Simulation of Motion}
\begin{figure}[!ht]
    \centering
    \includegraphics[width=.45\textwidth]{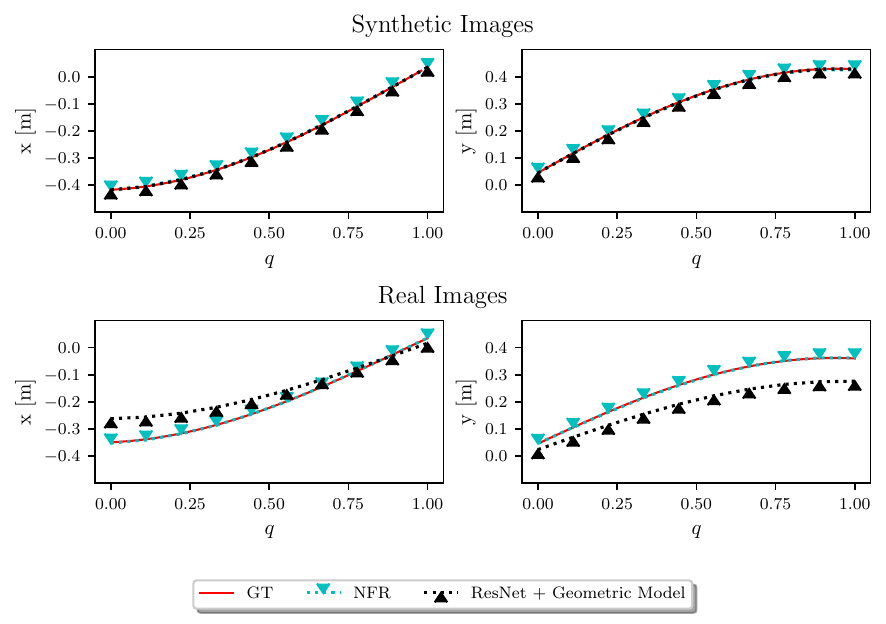}
    \caption{Predicted handle positions; latent codes inferred from synthetic images (top) and real images (bottom)}
    \label{fig:handle_prediction}
\end{figure}

Using a latent code $\mathbf{z}$ obtained from the synthetic dataset we are able to simulate the whole range of motion by generating new latent codes for arbitrary articulations $q \in [0, 1]$. For each generated latent code the predicted handle position is shown in Fig.~\ref{fig:handle_prediction}. With a traditional image encoder we are not able to compute handle positions for arbitrary articulations $q$ directly. Thus, based on the image encoding of the corresponding instance we estimated only the current articulation and the position of the hinges. Additionally, we predict the position of the handle and hinge joint for different articulations based on the explicit geometric model we provided for comparison.

Both approaches perform well in predicting the handle positions for arbitrary articulations on synthetic images, but if we compare their predictive performance on real objects our approach outperforms the baseline, which diverges from the true path (Fig.~\ref{fig:handle_prediction}). Here we are using the data from $\mathcal{D}_\text{REAL}$. 
Those images and the corresponding camera parameters are drawn from a different distribution than the one present in the synthetic dataset. 
Since our approach minimizes the reconstruction loss it is able to generalize to this out-of-distribution scenario.

\subsection{Motion Planning}
Last, we describe the integration of all parts for manipulation planning in simulation and on a real robot.

\subsubsection{Simulation}  Given only a small set of images from different viewpoints, we are able to estimate the current position of all keypoints and to simulate their movement during interaction with the robot. Only those keypoint predictions are used during trajectory optimization. After planning, we check that the handle is grasped correctly and that the motion does not violate the geometric constraints of the object.

\subsubsection{Real Robot}
For manipulation planning on a real robot, we take ten images from different viewpoints. Based on the latent code which minimizes the image loss we predict ten waypoints to formulate and solve the corresponding trajectory optimization (see Sec. \ref{sec:planning}). Finally, we execute the plan using a position-based controller.      

Thus, even without an explicit kinematic model of the perceived object, the robot is able to perform the desired object manipulation as shown in Fig. \ref{fig:pipeline} 
and in the accompanying video by forward simulating the motion.


\begin{figure}
\centering
\includegraphics[width=.16\textwidth]{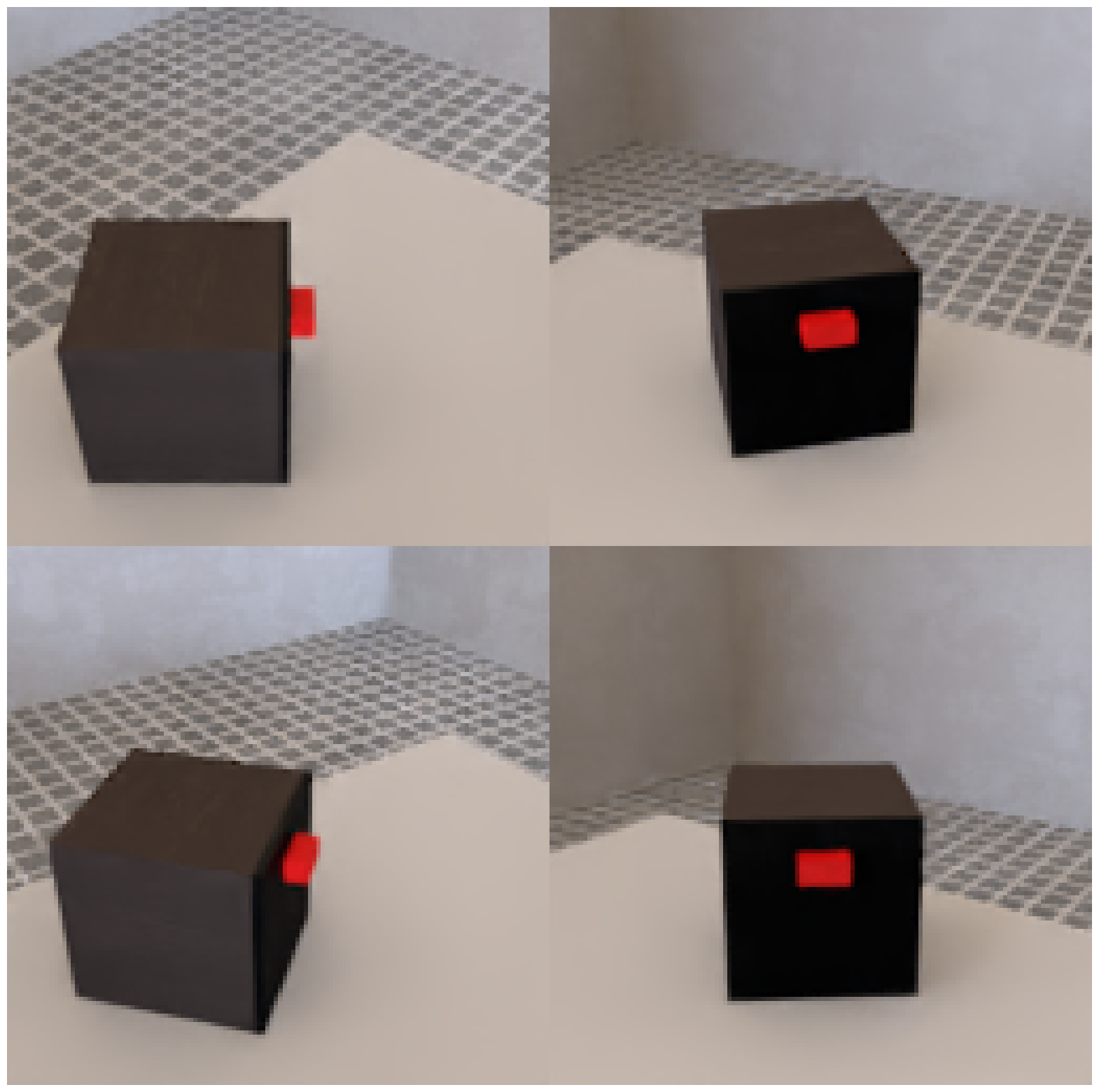}\hfill
\includegraphics[width=.16\textwidth]{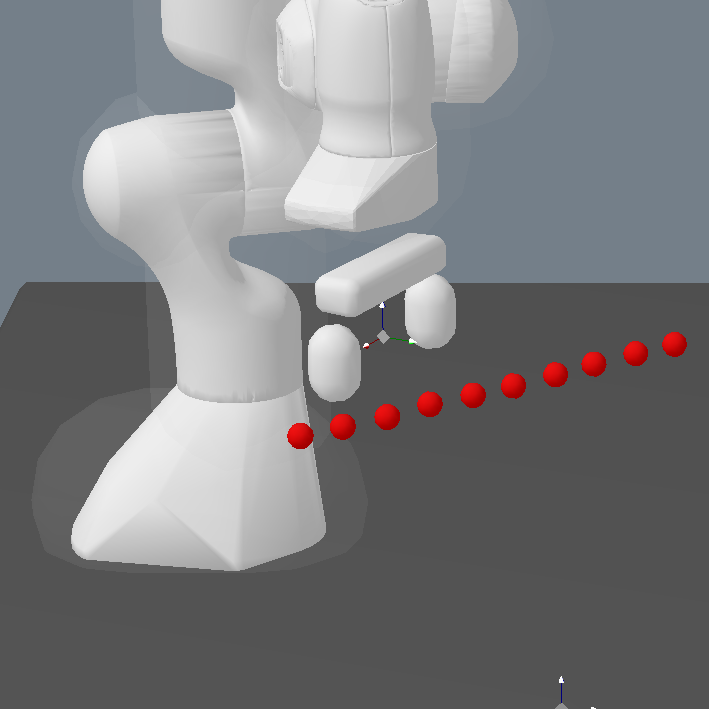}\hfill
\includegraphics[width=.16\textwidth]{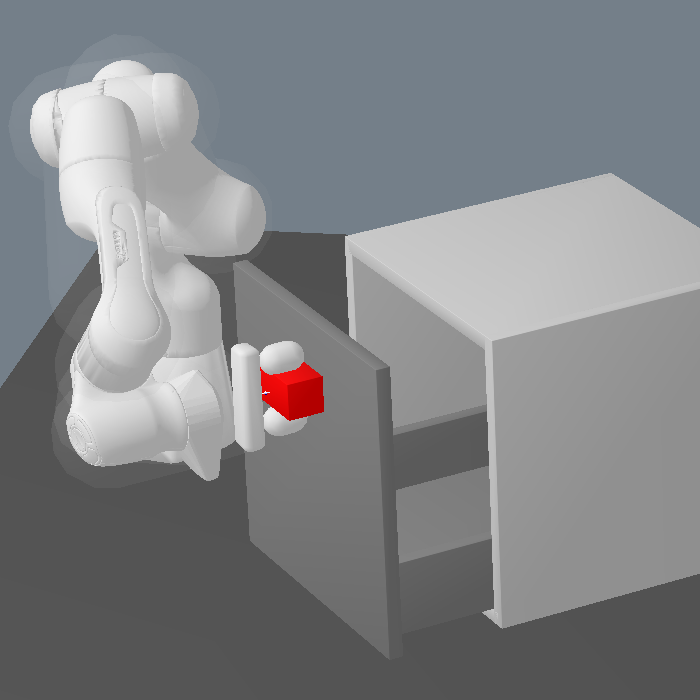}
\caption{Interaction with a drawer: Four input images (left); Predicted handle positions (middle); final interaction in simulation (right).}
\label{fig:drawer}
\end{figure}

\subsection{Generalization to Different Object Categories}
Our approach generalizes to different object categories. We trained a different model to manipulate drawers (Fig.~\ref{fig:drawer}).  
Note that objects of this class impose a different movement constraint compared to the closets.
With our method we can predict the handle positions for the entire range of motion and perform manipulation planning for drawers as well.

\section{CONCLUSION\label{sec:conclusion}}
In this work, we have proposed a method for finding implicit representations of articulated objects by minimizing the image loss between observed images and rendered images. As we have shown this approach is robust to out-of-distribution scenarios and generalizes to real images and previously unobserved camera parameters. The structured latent code enables motion planning by predicting keypoint position through forward simulating the motion of observed objects. Finally, we demonstrated manipulation planning in simulation and on a real robot.

A current limitation is that we trained separate models for different object classes (e.g., closets and drawers). As future work we would address this limitation by training a single general model with data of multiple diverse objects. Furthermore, in this work we considered only objects with a single joint. How our approach scales to complex objects with multiple joints is another interesting direction for further research.   

\addtolength{\textheight}{-1cm}   



%
%
%
%
%



\bibliographystyle{IEEEtran}
\bibliography{IEEEabrv,mybibfile}

\end{document}